# MOBILE KEYBOARD INPUT DECODING WITH FINITE-STATE TRANSDUCERS

*Tom Ouyang, David Rybach, Françoise Beaufays, Michael Riley*

Google

{`ouyang,rybach,fsb,riley`}@google.com

## ABSTRACT

We propose a finite-state transducer (FST) representation for the models used to decode keyboard inputs on mobile devices. Drawing from learnings from the field of speech recognition, we describe a decoding framework that can satisfy the strict memory and latency constraints of keyboard input. We extend this framework to support functionalities typically not present in speech recognition, such as literal decoding, autocorrections, word completions, and next word predictions.

We describe the general framework of what we call for short the keyboard "FST decoder" as well as the implementation details that are new compared to a speech FST decoder. We demonstrate that the FST decoder enables new UX features such as post-corrections. Finally, we sketch how this decoder can support advanced features such as personalization and contextualization.

## 1. INTRODUCTION

With the fast-growing penetration of mobile devices in every aspect of modern life, offering an efficient and pleasant mobile input experience has recently become a topic of interest to researchers and technology providers. Speech recognition for example has flourished in the last few years, mostly fueled by the need for convenient mobile input methods [1]. Likewise, handwriting recognition has gained more traction, especially in languages with complex scripts such as Chinese and Indic languages [2].

Keyboard input has received relatively less attention from the research community, even though it remains a primary input method as it is often considered, correctly or not, as the most convenient way to compose text on a mobile device. At first glance, keyboard input may seem trivial, but soft keyboards have long surpassed in capabilities the hardware keyboards used on laptops and desktops. They compensate with a rich feature set for the difficulty of typing on a small screen, they can generalize to a broad variety of languages, and being implemented on smart devices, they hold the promise of leveraging rich information about the user and their environment. An FST decoder, with its mathematical formalism and principled implementation, seems a natural choice to power such features.

The use of FSTs in the context of keyboard input is not totally new: a report by Klarlund and Riley suggested using FSTs to disambiguate entries on a hardware cluster keyboard, but this was before the invention of smart phones, so the issue of decoding (soft) noisy input sequences was not addressed [3]. FSTs have also been used for similar tasks like spelling correction [4], though not in the context of mobile input. The present paper contains to our best knowledge the very first description of an FST decoder for mobile keyboard input with production- level constraints.

While there have been many mobile keyboard products targeted to smartphones (e.g., Shapewriter [5], Swype, Swiftkey, etc.), the published literature and availability of datasets in this area is very limited. Where relevant, we will compare the FST decoder to a prior in-house implementation using a time-synchronous token-passing decoder model. General background on FST decoding for speech can be found in [6].

The rest of this paper contains a summary of major keyboard features and associated terminology (Section 2), a detailed description of the various keyboard transducers (Section 3) and decoder implementation (Section 4), and extensions to more advanced features (Section 5).

## 2. KEYBOARD FEATURES AND TERMINOLOGY

A mobile keyboard must above all be reliable and fast. For these reasons, all run-time processing happens on device. Latency constraints are tight: a key press is expected to produce visible feedback within about 20 msec. RAM and CPU usage must be kept under strict control to prevent the keyboard from being evicted by concurrent processes and to protect the device's battery life. Memory is also limited: as in embedded speech recognition, keyboard language models should not exceed 5 to 10 Mb, which typically allows them to model a couple hundred thousand words at most.

The main function of a soft keyboard is to decode touch inputs into words and sentences, just like a speech recognizer would decode input waveforms. Advanced keyboards support "tap typing", where users tap the keys corresponding to the characters of a word, and "gesture typing", where they swipe their finger across the keyboard layout. Users can switch freely from one input mode to the other within a sentence.

In tap mode, when the user enters a white space after a character sequence, the composed word is said to be "committed". In gesture mode, this happens when they lift their finger at the end of the swipe. Because of "fat finger" errors, the user may not tap the exact keys they intended. The sequence of characters indicated by the "bounding boxes" of the keys they actually pressed is called the "literal" decoding of the word. Recognition candidates with high combined spatial and language model scores are typically shown as "suggestions" on the "suggestion strip" right above the keyboard layout. When the score of the best suggestion exceeds that of the literal by some margin, the literal is replaced with the suggestion as an "autocorrection". The same mechanism is used to correct user typos. Literal decoding also allows users to enter words that are out-of-vocabulary (OOV) to the language model. Balancing OOV recognition and autocorrections is tricky. Gesture input typically does not allow OOV recognition and does not have a concept of literal decoding. It is more similar to speech recognition than tap input.

While a user starts entering a word key by key, word "comple-tions" are proposed that they can click on. When they commit a word, "next word predictions" are proposed instead. These two features help users enter fewer keystrokes and type faster. Completions are typically not offered in gesture mode as clicking on them would require the user to lift their finger which could signal they want to commit the word composed so far instead.

Of course, a soft keyboard also offers character deletions (backspace), and repositioning of the cursor at any time.

While gesture input is easily 10 to 20% faster than tap input [7], relatively few users rely primarily on gesture input.

## 3. KEYBOARD TRANSDUCERS

A tapped input consists of a time series of touch points, **x**, that encodes the coordinates of the user's key presses. For gesture input, the input trajectory is sampled, e.g. every 100 milliseconds, to provide a similar time series. The task of the decoder is to find the word sequence **w** that best matches the input sequence **x**.

### 3.1. Key Context Dependency and Spatial Model

The equivalent of speech phonemes in the keyboard world is the set of keys offered in the layout. Accordingly, a spatial model is used to provide a probability distribution over these units. Note that the spatial model does not resolve fully the written language: For example, the letters "é" and "è" in a French keyboard are typically obtained by long-pressing "e" and choosing from a small pop-up menu. All three letters have the same spatial score. This confusability is mitigated by imposing context dependency constraints and enforcing a strong language model.

Just like acoustic context dependency is encoded in speech with a $C$ transducer, we implemented spatial context dependency in keyboard, and chose to do so with a bi-key model. Accordingly, the arcs of the transducer represent $a\_b : b$ transitions, where $a\_b$ means the key $b$ with $a$ as left context (previous key). This places keyboard somewhere in between our server-based speech recognizer which relies on triphone models and our embedded recognizer which uses monophones [8].

The spatial model for tap input is typically a Gaussian distribution centered on each key center. Gesture inputs instead are often modeled with the so-called "minimum-jerk model" that imposes smoothness maximization constraints on the input trajectory [9]. Alternatively, a recurrent neural network model can be used [10].

### 3.2. Lexicon

The lexicon transducer $L$ for the keyboard decoder is a simple key to word mapping, like a speech grapheme lexicon. Some keys like the apostrophe may be made optional, allowing users to type "Ive" for "I've". Similarly, repeated keys can be defined optional, e.g. the second "o" in "Google". The closure, which allows transducing sequences of words, is implemented with a space symbol for finger lift-up or taps on the space bar between words. If this space symbol is optional, the decoder can correct missing space taps between words or decode multi-word gestures. An example of a lexicon FST is shown in Figure 1.

### 3.3. Language Model

Similar to the language models in embedded speech recognition systems, keyboard language models are typically low order n-grams over a limited vocabulary, e.g. 64K words [8].

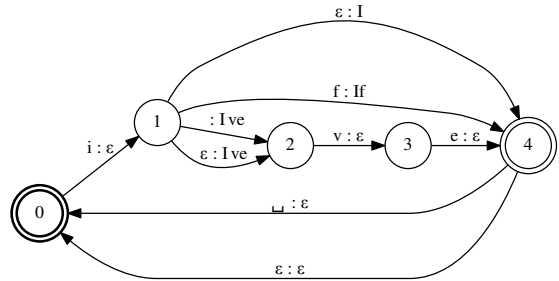

**Fig. 1**: Example of a lexicon FST for the words *I*, *I've* and *If* with optional space and optional apostrophe.

Because of features like suggestions, completions and predictions where the language model is more prominent than the spatial model, the language model should be carefully crafted. For example, the user who gestures "Google" may expect the keyboard to suggest "goggle" as an alternative, but not "gogle" or "gooogle", which chances are would be present in the training corpus as training corpora are often noisy. For this reason, keyboard language models are typically trained to a fixed vocabulary that has been hand-curated to eliminate misspellings, erroneous capitalizations, and other undesired artifacts.

With this, the decoder graph for keyboard is constructed using the composition of context, lexicon, and language transducers - the familiar $C \circ L \circ G$. For memory efficiency, we use the on-the-fly composition of $(C \circ L) \circ G$ using look-ahead composition filters [11].

Figure 2A shows a gesture input with its possible alignments to a simplistic, two-word decoder graph. Figure 2B shows the input sequence and its alignment for a similar tap sequence. The same decoder graph (Figure 2C) is shared for both types of input.

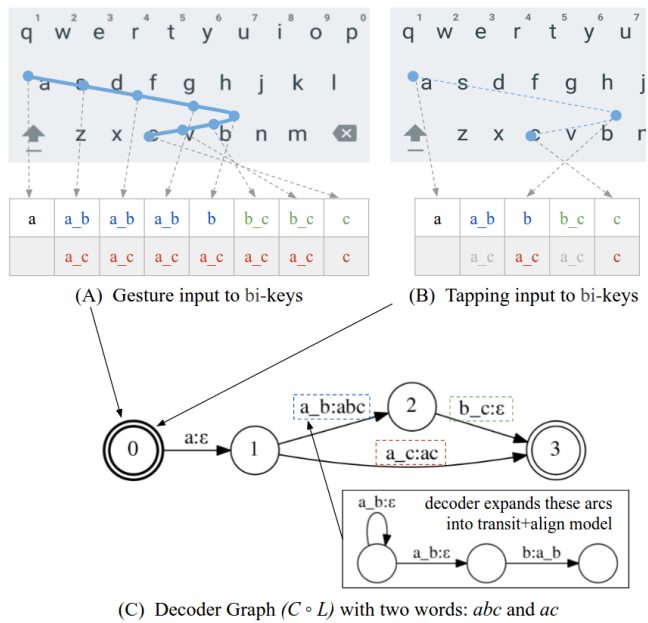

(A) Gesture input to bi-keys

(B) Tapping input to bi-keys

(C) Decoder Graph *(C ∘ L)* with two words: *abc* and *ac*

**Fig. 2**: A simple example of gesture (A) and tap input (B), with their state alignment to a 2-word toy FST (C) allowing only the words "abc" and "ac".

# 4. KEYBOARD DECODING WITH FSTS

## 4.1. Suggestions and Autocorrections

Text entry on a mobile touch-screen device is a very error-prone process, with per-letter error rates around 8-9% [12]. This is commonly known as the "fat finger" problem. One reason we don't constantly notice these errors in our everyday smartphone usage is because of the autocorrection functionality available on all modern keyboards. They work in the background, silently correcting our typos and misspellings.

*Substitutions*: One basic type of error is mistakenly substituting one letter for another (e.g., "tjis" instead of "this"). These include both mechanical "fat finger" errors, where one accidentally taps on a neighboring key and semantic spelling mistakes . The decoder's spatial model handles substitutions by producing a probability distribution over the set of possible keys for each frame/tap. This substitution model takes into account the proximity between the touch point and the key.

*Deletions*: Deletions occur when the user does not produce any input tap for one or more letters in the intended word (e.g., "farenheit" instead of "fahrenheit"). To handle these errors, the decoder allows to skip one (non-epsilon) arc per frame in the lexicon FST without any input. The weight for these extra epsilon-transitions should represent the probability of omitting the letter in the given context.

*Insertions*: The spatial model also allows taps to be treated as extraneous inputs due to accidental touches or spelling errors (e.g., "truely" instead of "truly"). These inputs are consumed without advancing the state in the decoder graph.

*Autocorrection*: No suggestion mechanism is perfect, so when the decoder finds a correction or prefix-completion candidate for the user's input, it must decide whether or not it is confident enough to trigger an autocorrection. If so, this candidate will automatically replace whatever was typed after the user enters the next separator (e.g., space or punctuation).

If the model is too conservative, it will leave many typos uncorrected - even though the decoder may have generated the correct candidate. If the model is too aggressive, it will start making false autocorrections away from correctly typed words, sometimes without the user noticing. This can be very frustrating, since it can drastically change the meaning of the text. For example, imagine the keyboard autocorrecting "you're far" to "you're fat" (since the latter may be preferred by the language model and on a QWERTY layout the R is adjacent to the T).

To moderate its autocorrection aggressiveness, the decoder evaluates the tapped characters (known as the "literal") using a character-level model. This allows it to calculate a probabilistic score for OOV words, and rank it against the in- vocabulary corrections. It also applies an additional cost for correcting away from an already valid word, like the "far" to "fat" example. This reflects the higher user annoyance whenever the keyboard "auto-corrupts" an already correctly-typed word.

## 4.2. Gesture Typing

For gesture typing, the interaction model typically assumes each sliding stroke corresponds to a single word, and space between words are inserted automatically if omitted. This type of input can be especially challenging because the finger often slides past many keys while in-transit to the actual intended letter. For example, on a typical QWERTY keyboard layout, the words "pit", "pot", "put" all have the same canonical gesture pattern (a straight line from P to T).

The same FST decoding framework used for tap typing also works naturally for gesture typing. Here, each input frame is interpreted as either in-transit to a key (e.g., the "$a\_b$" label in Figure 2A) or aligned to a key (e.g., the "$b$" label). The in-transit label is analogous to an insertion in tap input, where the input can be consumed in a self-loop without advancing the search state.

## 4.3. Literals Decoding

In order to include the literal decoding of tap input in the hypotheses returned by the keyboard decoder, the decoder graph must contain paths for arbitrary key sequences. The lexicon includes for each key an entry with a corresponding "literal word" on the output side. In contrast to regular words, the lexicon does not permit a space character between these literal word symbols. The input symbols for the literals words in the lexicon differ from regular key symbols, such that the spatial model can assign only that "literal key" symbol a non zero probability when it was actually tapped.

The LM FST contains a subgraph for the literal words grammar, which assigns weights to sequences of literal words. It is connected to the unigram state of the model. The grammar also inserts a marker token at the end of the literal sequence. This marker enforces a space tap to separate it from the next word (or the end of the input).

In a post-processing step after decoding, sequences of literal words in the decoder result are combined to form word symbols in the output.

## 4.4. Word Completions and Predictions

For word completion prediction, the decoder computes the most likely extensions for each of the currently hypothesized word prefixes in their respective sentence contexts. The decoder keeps as hypotheses a set of states in the decoder graph along with their score (and a traceback pointer). The states in the decoder graph correspond to a tuple of states in the lexicon, the LM FST, and the composition filters.

From a lexicon state, we can compute the set of reachable word (output) labels. In fact, the label look-ahead composition filter requires that information as well and it is therefore available as precomputed interval set [11]. Due to minimization of the lexicon FST and label pushing, the output labels may appear already on an arc before the last key of a word (for example, see states 2 and 3 in Figure 1). Therefore, the word completion procedure must not only look forward in the graph but also backwards in the traceback data and the label look-ahead composition filter state.

The reachable words are looked up in the LM at the state obtained from the corresponding hypothesis' decoder graph state. Note that parts of the LM score may already be incorporated in the hypothesis score, due to on-the-fly weight pushing. The LM probability is combined with the hypothesis score to obtain a score for the word completion. We collect only a small set of the $n$-best unique words.

The next word is predicted from the current best (partial) sentence. We find the highest order n-gram state for the sentence prefix and collect the set of $n$-best unique output labels from that state, including paths over backoff transitions. The arcs in the LM FST are sorted by label, not by score, for efficient composition. Hence, the next word prediction must traverse all arcs of a state. Caching must be used to decrease the high computational cost for states with large outdegree, in particular the unigram state.

The integration of dynamic models (cf. Section 5.1) and the prediction of next words and word completions from the dynamically added vocabulary requires additional lookups in these models.

## 4.5. Post-Corrections

It can be difficult for the language model to determine if a word makes sense without seeing what comes after it. For example, there is nothing wrong with the word "food" at the start of a sentence, but if the user goes on to type "food luck", there is a good chance that she actually meant "good luck" (especially since on the QWERTY layout the F key is adjacent to the G key).

However, traditional smartphone keyboards treat the space bar as a hard commit, with no ability to change its mind about previously typed words. This is analogous to the early state of speech recognition, when most systems operated on isolated words. By adopting a modern continuous recognition architecture, the FST decoder enables a streaming interaction model where the keyboard can adjust its interpretation of previous words based on new evidence.

There are also new user interaction challenges in allowing the keyboard to modify previously committed words. Users may be initially surprised by this unexpected behavior, and false-post-corrections could be especially frustrating (and time consuming to fix). One approach to mitigating these risks is to make post-corrections more conservative. E.g., by permitting them only within a small temporal window of the most recently typed word, and only when the decoder has a very high confidence in the replacement.

## 5. FST DECODER AND COMPLEX FEATURE NEEDS

Perhaps one of the most powerful aspects of the FST framework for keyboard input is the ease with which dynamic models can be exploited for personalization and contextualization. This is especially important with on-device decoding, where personalization can compensate for the limited size of the models: one rarely needs more than 100K words to express themselves in a given language, but they may need the *right* 100K words.

### 5.1. Dynamic Models

Dynamic models can be used to accumulate n-grams the user has previously typed, or information such as their contact list, or other contextual information. These models are constantly updated, so it would be challenging and costly to perform such updates directly on the decoder graph. Instead, they are incorporated in the decoding process using an on-the-fly lattice rescoring technique, similar to the n-gram biasing approach described in [13].

The vocabulary of the dynamic LM can contain words which are not covered by the main LM and are therefore not part of the lexicon FST. We integrate these OOV words in the decoder graph by splicing a character to word transducer to the LM FST, connected to the unigram state. This FST has "character words" on the input side and regular words on the output side. The lexicon FST has arcs that map these character word sequences to the corresponding (bi-) key sequences.

## 6. EVALUATION

We evaluate performance on a dataset of tapped and gestured sentences (over 3000 words total for each input modality). In the data collection study, participants tapped and gestured English sentences on a smartphone QWERTY keyboard. The prompts were sentences sampled from transcribed speech interactions. Since we wanted to collect natural errors to test the correction capability of the decoder, participants were asked to type quickly and not worry about fixing any mistakes.

The results below show the performance of three keyboard decoders. A Baseline system without label look-ahead or post-correction (p.c.), the proposed FST decoder without p.c., and the same FST decoder with p.c. enabled.

| Decoder | Tapping WER | Gesture WER |
|---|---|---|
| Baseline | 6.31% | 11.90% |
| FST | 5.78% | 11.51% |
| FST w/ p.c. | **5.07%** | **9.20%** |

**Table 1**: Word error rates between different keyboard decoders.

| Prompt | *it meant a whole lot of something* |
|---|---|
| Baseline | it meant a while lot of something |
| FST | it meant a while lot of something |
| FST w/ p.c. | it meant a whole lot of something |

**Table 2**: One example sentence that triggered post-correction.

Table 1 shows that the proposed FST decoder is able to outperform the baseline for both tap and gesture input. The relative reduction in WER becomes even greater when post-correction is enabled. The example from Table 2 shows this in action. By taking into account the following word, the p.c. decoder was able to revise the erroneous "a **while** lot" to the correct "a **whole** lot".

## 7. CONCLUSION

We showed how the FST framework developed over the last decade for speech recognition can be ported to the world of mobile keyboard input, from the basic decoding of tapped key sequences to more advanced features such as next word prediction. Our experiments so far show that an FST decoder brings strong accuracy advantages over a more traditional decoder, that it can elegantly scale to the challenges of internationalization, and that it can readily exploit on-device data to personalize the user's typing experience.